\documentclass[runningheads,a4paper]{llncs}  
\usepackage[T1]{fontenc}
\usepackage{lmodern}                          %
\usepackage{graphicx}
\usepackage{amsmath}
\usepackage{amssymb}
\usepackage{float}
\usepackage{booktabs}   
\usepackage{multirow}  
\usepackage{siunitx}    
\usepackage{placeins}   
\usepackage{subcaption}
\usepackage{microtype} 
\begin{document}

\setlength{\abovedisplayskip}{5pt plus 1pt minus 2pt}
\setlength{\belowdisplayskip}{5pt plus 1pt minus 2pt}
\setlength{\abovedisplayshortskip}{2pt plus 1pt}
\setlength{\belowdisplayshortskip}{5pt plus 1pt minus 2pt}
\setlength{\textfloatsep}{10pt plus 2pt minus 4pt}
\setlength{\floatsep}{8pt plus 2pt minus 2pt}
\setlength{\intextsep}{8pt plus 2pt minus 2pt}
\setlength{\abovecaptionskip}{4pt}
\setlength{\belowcaptionskip}{0pt}

\title{Adversarial Attacks on Online Handwriting using Salience-based Temporal Editing}
\titlerunning{Adversarial Attacks on Online Handwriting}

\author{Yataro Tamura \and
Brian Kenji Iwana\orcidID{0000-0002-5146-6818} \and
Jiseok Lee\orcidID{0009-0004-2718-2227}}

\authorrunning{Y. Tamura et al.}

\institute{Kyushu University, Fukuoka, Japan\\
\email{\{yataro.tamura, jiseok.lee\}@human.ait.kyushu-u.ac.jp}\\
\email{iwana@ait.kyushu-u.ac.jp}}

\maketitle

\begin{abstract}
Deep learning models for online handwriting recognition have been shown effective and are increasingly deployed in practical applications. However, their vulnerability to adversarial attacks is still a challenge. Existing adversarial methods are predominantly designed for image-based inputs and typically rely on additive spatial perturbations. When applied to online handwriting, which is inherently represented as a time series of pen trajectories, such perturbations often introduce high-frequency jitter and visibly unnatural stroke artifacts. In this work, we propose a novel adversarial attack framework for online handwriting recognition based on salience-guided temporal editing. Instead of adding noise, the proposed method generates adversarial examples by inserting and deleting points at time steps selected according to temporal salience, preserving the shape and smoothness of the original handwriting. Temporal salience is estimated using gradient-based activation mapping, which guides edits toward time steps that strongly support the original class prediction. We evaluate the proposed approach on the Unipen and CASIA-OLHWDB datasets under both white-box and one-shot black-box attack settings. Experimental results demonstrate that while conventional image-based attacks achieve strong white-box performance, they exhibit poor transferability across models. In contrast, the proposed temporal editing attack achieves stronger one-shot black-box transferability while preserving the visual structure of the handwriting. These results indicate that temporal editing is a relevant threat model for online handwriting recognition, particularly in one-shot black-box transfer settings.

\end{abstract}

\section{Introduction}

Online handwriting recognition differs from offline handwriting recognition in that the input is a sequence of pen-tip coordinates rather than a raster image.
Modern online handwriting recognizers build on sequence models such as Recurrent Neural Networks (RNN)~\cite{rumelhart1985learning}, Temporal Convolutional Neural Networks (CNN)~\cite{lang1990time}, and Transformers~\cite{vaswani2017attention}, and have achieved high accuracy on this input modality~\cite{Graves2009TPAMIHandwriting,iwana2020time,ghosh2022advances,liu2025col,qu2025end}.
However, adversarial robustness for online handwriting remains less explored than for image-based recognition models.

As these systems are deployed in real-world applications, concerns about their reliability and security have become more pronounced.
In particular, adversarial attacks are a widely known weakness of neural networks~\cite{xu2020adversarial,costa2024deep}. 
Adversarial attacks work by adding small perturbations to the input with the aim of causing misrecognition by the models.

However, most adversarial attacks were originally developed for image-based classifiers with perturbations constrained by pixel-wise $L_p$ norms.
For example, various gradient-based attack methods have been proposed, such as Fast Gradient Sign Method (FGSM)~\cite{Goodfellow2015}, Projected Gradient Descent (PGD)~\cite{Madry2018}, and the Carlini-Wagner (CW) attack~\cite{CarliniWagner2017}. 
These methods are designed to degrade recognition performance while remaining imperceptible in the image domain.
In comparison, attacks on time series are much less explored, and most adversarial attacks on time series are adapted from image-based methods~\cite{karim2020adversarial,Fawaz2019AdversarialTSC}.

\begin{figure}[t]
    \centering
    \includegraphics[width=1.0\textwidth]{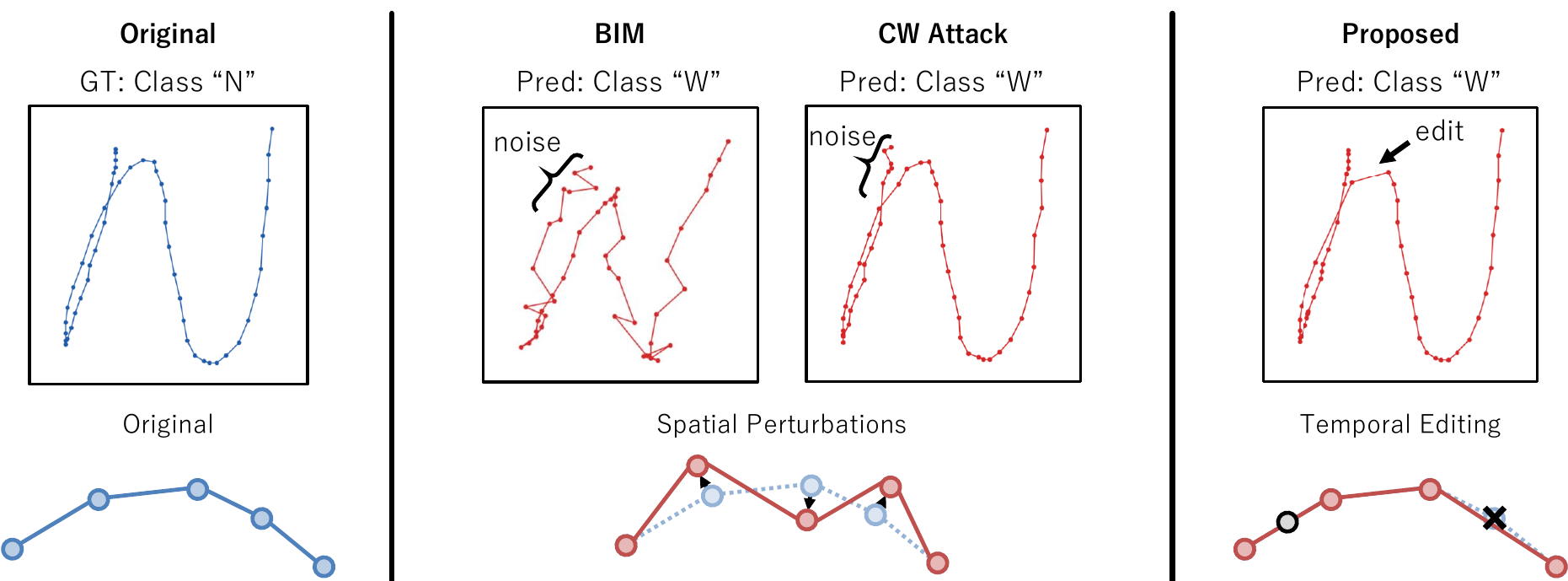}
    \caption{Comparison between adversarial noise via perturbations and the proposed temporal editing. Adversarial attacks such as BIM and CW use spatial perturbations to generate adversarial examples. The proposed method adds and removes time steps for more realistic characters.}
    \label{fig:example}
\end{figure}

The online handwriting data considered in this study is represented as a sequence of two-dimensional pen-tip coordinates, where each time step $t$ contains a coordinate vector $\mathbf{x}_t \in \mathbb{R}^2$.
In the image domain, small perturbations can appear as inconspicuous noise due to the texture present in natural images.
However, directly applying image domain adversarial noise to time series data frequently results in abrupt temporal fluctuations and high-frequency artifacts, which create unnatural deformations in online handwriting trajectories, as illustrated in Fig.~\ref{fig:example}.
This indicates that even when perturbations satisfy small $L_p$ norms, the resulting time series may violate the kinematic smoothness and structural regularities inherent to human handwriting.
In other words, the implicit assumption in the image domain that “small noise is difficult for humans to perceive” is not necessarily valid for time series data.
This discrepancy highlights a fundamental limitation in current adversarial attacks on time series.

Beyond naturalness issues, most gradient-based adversarial attacks operate under a white-box assumption, requiring full access to model parameters and gradients.
In practice, however, deployed handwriting recognition systems are often accessible only through prediction interfaces, making black-box adversarial attacks a more realistic threat model.
Commercial handwriting recognition and authentication services typically expose only final recognition outputs, rendering white-box assumptions impractical in many real-world scenarios.
Therefore, evaluating robustness under black-box conditions is essential for understanding the true vulnerability of online handwriting recognition systems.

To address these challenges, we propose Adversarial Iterative Temporal Editing (AITE), a novel adversarial attack method that generates adversarial examples through discrete editing operations in the time dimension rather than additive noise in the spatial dimensions. 
Specifically, AITE iteratively performs element-wise insertion and deletion on the input sequence based on the discriminative importance, or salience, of individual time steps.
Salience is estimated using Gradient-weighted Class Activation Maps (Grad-CAM)~\cite{Selvaraju2017GradCAM}, allowing the attack to target temporally critical regions while preserving the overall geometric structure and kinematic smoothness characteristic of natural handwriting trajectories.

The contributions of this study are summarized as follows:
\begin{itemize}
    \item We highlight the inadequacy of spatial perturbation-based adversarial noise for online handwriting.
    \item We propose a novel adversarial attack framework based on salience-guided temporal editing.
    \item The proposed method is evaluated under both white-box and one-shot black-box attack settings on the Unipen~\cite{Guyon1994UNIPEN} and CASIA-OLHWDB~\cite{Liu2011CASIA} online handwritten character datasets. Experimental results demonstrate that temporal editing achieves stronger visual similarity than conventional spatial perturbation-based noise attacks.
    \item The black-box attack settings demonstrate that the salience-based editing is effective at attacking unseen models.
\end{itemize}
Furthermore, for reproducibility, we will release the source code at GitHub\footnote{\url{https://github.com/yataro0117/AITE}}.

\section{Related Work}

\subsection{Adversarial Attacks}

Adversarial attacks were first extensively studied in the context of image classification, where small, carefully crafted perturbations can induce misclassification while remaining visually imperceptible~\cite{costa2024deep}. 
The vulnerability of deep neural networks to adversarial perturbations was first systematically demonstrated by Szegedy et al.~\cite{szegedy2013intriguing}, who formulated adversarial example generation as an optimization problem that searches for minimal input perturbations capable of changing a model’s prediction.
To reduce the computational cost of this approach, Goodfellow et al.~\cite{Goodfellow2015} proposed FGSM, which approximates the optimization by applying a single gradient step in the direction of the loss.
Since then, many methods, such as Basic Iterative Method (BIM)~\cite{Kurakin2017AdversarialExamples}, PGD~\cite{Madry2018}, and the CW attack~\cite{CarliniWagner2017}, have improved upon gradient-based adversarial attacks through iterative updates and regularized perturbation objectives.
A separate line of work targets the black-box transferability of adversarial examples between models.
MI-FGSM~\cite{Dong2018MIFGSM} accumulates a momentum term over the iterative input gradients to stabilize the update direction, while NI-FGSM~\cite{Lin2020NIFGSM} extends MI-FGSM with Nesterov accelerated gradients by evaluating the gradient at a look-ahead point.
TI-FGSM~\cite{Dong2019TIFGSM} convolves the input gradient with a smoothing kernel before each update so that perturbations become approximately translation-invariant and transfer better across architectures.
While these methods form the foundation of adversarial attack research, they were all originally developed for image-based classifiers and rely on additive, pixel-wise perturbations.

\subsection{Adversarial Attacks on Time Series}
Compared to image recognition, research on adversarial attacks against time series data is an emerging area.
Carlini et al.~\cite{Carlini2016HiddenCommands} demonstrated that hidden commands can be embedded into automatic speech recognition systems. 
Fawaz et al.~\cite{Fawaz2019AdversarialTSC} evaluated FGSM and BIM on time series classification datasets and reported the vulnerability of deep learning--based time series classifiers.
Karim et al.~\cite{karim2020adversarial} proposed an Adversarial Transformation Network (ATN) using a distilled model to act as a surrogate to mimic the behavior of classical time series classification models.
In addition, there have been black-box models that have been proposed for time series classification~\cite{rathore2020untargeted,yang2022tsadv,ding2023black}.
However, these methods still apply perturbations which may not be appropriate for all time series. 
Furthermore, many of these methods require feedback from the attacked model~\cite{zheng2025blackboxbench}.
Developing attack methods that take into account structures specific to time series data (e.g., temporal dependency and sequence smoothness) remains an important research challenge~\cite{pialla2022smooth}.

\subsection{Adversarial Attacks on Handwriting Recognition}
Adversarial attacks on handwriting recognition have mainly been studied in the context of offline handwriting recognition, where adversarial attacks on offline handwriting have been shown to be effective on offline signature verification~\cite{hafemann2019characterizing,li2021black,guo2024white} and character recognition in various languages~\cite{jiang2022adversarial,shi2025generative,huynh2025adversarial}.

In contrast, research on adversarial attacks against online handwriting recognition is very limited.
Yamashita et al.~\cite{yamashita2024test} showed that spatial perturbation-based attacks work on online handwriting and proposed a defense based on time series transformations.
Lopresti and Raim~\cite{lopresti2005effectiveness} showed that a generative attack on online handwriting is possible.

\section{Adversarial Attacks}

\begin{figure}[t]
    \centering
    \includegraphics[width=1.0\textwidth]{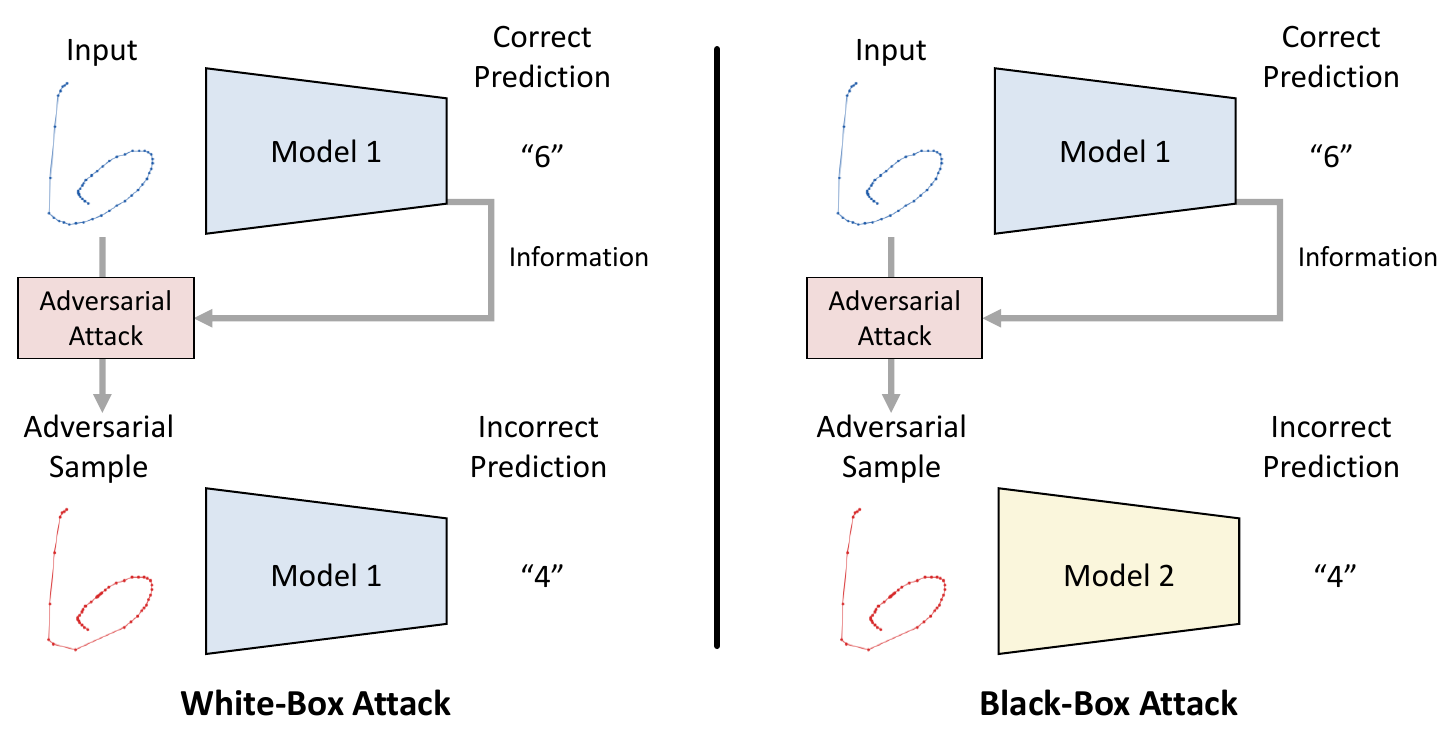}
    \caption{Example of transferability of adversarial attacks. In the white-box setting (left), the adversarial attack has full knowledge of the target model. In the black-box setting (right), the adversarial sample is generated against one model and is used to attack an unseen target model.}
    \label{fig:threat}
\end{figure}

\subsection{Threat Model}

We consider adversarial attacks against online handwriting recognition systems whose inputs are pen-tip trajectories of handwritten characters. In this study, each character is represented as a time series $\mathbf{X}$ with ground-truth label $y$:
\[
\mathbf{X} = (\mathbf{x}_1, \dots, \mathbf{x}_t, \dots, \mathbf{x}_T),
\]
where $T$ is the sequence length and $\mathbf{x}_t \in \mathbb{R}^d$ denotes the pen-tip coordinate vector at time step $t$. In our experiments, we use two-dimensional coordinate sequences, i.e., $d=2$. 
A recognition model \(f_\theta\) with trained parameters \(\theta\)
maps \(\mathbf{X}\) to class logits \(f_\theta(\mathbf{X}) \in \mathbb{R}^{C}\).
The predicted class label is
\[
    \hat{y} = \arg\max_{c} f_\theta(\mathbf{X})_c .
\]
The goal of an adversarial attack is to find an adversarial example
\(\mathbf{X}_{\mathrm{adv}}\) such that
\[
    \hat{y}_{\mathrm{adv}}
    = \arg\max_{c} f_\theta(\mathbf{X}_{\mathrm{adv}})_c,
    \quad
    \hat{y}_{\mathrm{adv}} \neq y,
\]
while the resulting handwriting remains visually plausible.

Adversarial attacks can be broadly categorized into white-box and black-box attacks, and we evaluate the proposed method under both settings.
In the white-box setting, the attacker has full access to the model architecture and parameters $\theta$ and the gradients of $f_\theta(\mathbf{X})$.
This allows the attacker to design $\mathbf{X}_\mathrm{adv}$ so that it directly exploits the model's decision boundary.
In the black-box setting, the attacker has no access to the model's parameters or gradients.
This setting reflects realistic deployment conditions, as malicious attackers would not have access to the model directly.

Furthermore, as illustrated in Fig.~\ref{fig:threat}, we consider one-shot transferable black-box attacks. 
In this setting, the adversarial attack method is permitted only a single attempt and receives no feedback from the target model. 
Transferability is particularly important for real-world scenarios, as deployed handwriting recognition systems typically expose only final predictions, preventing iterative probing or gradient access by an adversary~\cite{Papernot2016Transferability}.
Our black-box evaluation is therefore transfer-based and one-shot: adversarial samples are generated using a surrogate model and then submitted once to unseen target models.
We do not consider query-based black-box attacks, where the adversary adaptively updates the input using repeated feedback from the target model.

\subsection{Spatial Perturbation-based Adversarial Attacks}

Spatial perturbation-based adversarial attacks typically generate adversarial examples by adding small-magnitude noise to input representations.
As a representative example, FGSM~\cite{Goodfellow2015} creates an adversarial example $\mathbf{x}_{\mathrm{fgsm}}$ by:
\begin{equation}
\mathbf{X}_{\mathrm{fgsm}} = \mathbf{X} + \epsilon \cdot \mathrm{sign}\!\left(\nabla_\mathbf{X} L(f_\theta(\mathbf{X}), y)\right),
\label{eq:fgsm}
\end{equation}
where $L(f_\theta(\mathbf{X}), y)$ is the loss function between the model prediction for input $\mathbf{X}$ and the label $y$, $\nabla_\mathbf{X} L(f_\theta(\mathbf{X}), y)$ is the gradient of the loss with respect to the input, $\mathrm{sign}(\cdot)$ is the element-wise sign function, and $\epsilon$ is a hyperparameter controlling the perturbation magnitude. 
Iterative methods, such as BIM~\cite{Kurakin2017AdversarialExamples} and PGD~\cite{Madry2018}, repeat this process given constraints.

\section{Proposed Method}

We propose a temporal editing attack that generates adversarial examples by modifying the time series in the time dimension rather than adding spatial noise. 
The key idea is to induce misclassification through insertion and deletion of time steps while preserving the overall shape and smoothness of the handwriting.
As shown in Fig.~\ref{fig:example}, this design directly addresses the limitation of image-based adversarial perturbations, which often produce unnatural artifacts when applied to time series, including online handwriting.

\subsection{Temporal Editing}

In order to modify the time series in the time dimension, we define two basic editing operations, temporal insertion and temporal deletion.

\subsubsection{Temporal Insertion}

Given input time series $\mathbf{X}$ with $\mathbf{x}_t\in\mathbb{R}^d$, temporal insertion adds a new point between two adjacent points $(\mathbf{x}_i,\mathbf{x}_{i+1})$ with $i\in\{1,\ldots,T-1\}$. 
Interpolation is used to preserve local continuity and the new sequence becomes:
\begin{equation}
    \label{eq:insertion}
    \mathbf{X}_\mathrm{ins}=(\mathbf{x}_1,\dots,\mathbf{x}_{i},\frac{\mathbf{x}_i + \mathbf{x}_{i+1}}{2},\mathbf{x}_{i+1}\dots,\mathbf{x}_T),
\end{equation}
increasing the number of time steps by one.

\subsubsection{Temporal Deletion}
Temporal deletion removes one time step from the sequence. 
Since removing an endpoint may severely distort the overall shape, we exclude endpoints from deletion candidates. 
Given a selected time index $j$, deletion removes the $j$-th element from $\mathbf{X}$ to form:
\begin{equation}
    \label{eq:deletion}
    \mathbf{X}_\mathrm{del}=(\mathbf{x}_1,\dots,\mathbf{x}_{j-1}, \mathbf{x}_{j+1},\dots,\mathbf{x}_T).
\end{equation}
The resulting $\mathbf{X}_\mathrm{del}$ has one fewer time step than the original $\mathbf{X}$.

\subsection{Adversarial Iterative Temporal Editing (AITE)}

To create adversarial examples, an iterative approach is taken.
In one step of AITE, we select an insertion position $i$ and a deletion position $j$, and apply the edit to the current sequence to generate a new sequence. 
By applying insertion and deletion in equal numbers, the sequence length is always kept at $T$ while locally modifying only the trajectory shape.
This editing step is repeated until the predicted label no longer matches the true label $y$ (attack success), or until the maximum number of iterations $K$ is reached.

The following constraints are imposed to ensure stable editing:
\begin{itemize}
    \item In each iteration, we perform one insertion and one deletion so that the sequence length remains $T$. This allows the generated adversarial examples to be used for any model without interpolation or similar adjustments.
    \item The element inserted in the same iteration is not allowed to be deleted. This constraint prevents iterations without changes.
    \item Endpoints are excluded from deletion to prevent excessive distortion of the trajectory.
\end{itemize}

\subsection{Salience Estimation for Temporal Editing}

In order to select the locations of the inserted and deleted elements, we estimate temporal salience using Grad-CAM~\cite{Selvaraju2017GradCAM}.
Grad-CAM is a post-hoc explainability method that highlights regions of an input that are most influential for a model's prediction by leveraging gradient information from intermediate feature representations.
Originally proposed for images, in this work, we adapt Grad-CAM to be used for time series by estimating the element-wise salience instead of pixel-wise salience.

Let $\mathbf{A}\in\mathbb{R}^{M\times T'}$ denote an intermediate temporal feature map, where $M$ is the number of channels and $T'$ is the temporal length of the feature map.
We write $A^k\in\mathbb{R}^{T'}$ for the $k$-th channel of $\mathbf{A}$.
Because temporal pooling layers in the CNN can change the temporal resolution, $T'$ may differ from the input length $T$.
Let $z_c=f_\theta(\mathbf{X})_c$ denote the pre-softmax logit for class $c$.
In Grad-CAM, channel weights are defined as:
\begin{equation}
\label{eq:Grad-CAM-alpha} 
\alpha_k^{(c)} = \frac{1}{T'}\sum_{\tau=1}^{T'}\frac{\partial z_c}{\partial A_\tau^k}.
\end{equation}

The feature-level temporal contribution is then obtained as:
\begin{equation}
    \tilde{s}_\tau^{(c)}=\sum_{k=1}^{M}\alpha_k^{(c)} A_\tau^k,\quad\tau=1,\ldots,T'.
\end{equation}
Unlike the original Grad-CAM formulation, AITE does not apply the ReLU non-linearity to $\tilde{s}_\tau^{(c)}$, so that both positive and negative contributions to the original class are preserved for temporal edit selection.

Since $\tilde{\mathbf{s}}^{(c)}$ has the temporal resolution of the intermediate feature map, we resize it to the input length before selecting edit positions. Specifically, we linearly interpolate
\begin{equation}
\tilde{\mathbf{s}}^{(c)}=(\tilde{s}_1^{(c)},\ldots,\tilde{s}_{T'}^{(c)}),
\end{equation}
to length $T$, yielding the input-level salience sequence $\mathbf{S}^{(c)}=(s_1^{(c)},\ldots,s_T^{(c)}).$

When $T'=T$, this interpolation is the identity operation.
The input-level salience sequence $\mathbf{S}^{(c)}$ assigns a score to each original input time step and is used to rank temporal locations for editing.

\subsubsection{Temporal Insertion with Grad-CAM}
For temporal insertion, we choose the interval between two adjacent input time steps that has the lowest average salience:
\begin{equation}
\label{eq:insertion_grad}
i^\ast=\arg\min_{i\in\{1,\ldots,T-1\}}\frac{s_i^{(y)}+s_{i+1}^{(y)}}{2},
\end{equation}
where $y$ is the original class. We insert the midpoint of the selected interval $(\mathbf{x}_i,\mathbf{x}_{i+1})$ and update the sequence according to Eq.~\ref{eq:insertion}.

\subsubsection{Temporal Deletion with Grad-CAM}

For temporal deletion, after excluding endpoints, the deletion index is selected as:
\begin{equation}
\label{eq:deletion_grad}
j^\ast = \arg\max_{t\in\{2,\ldots,T-1\}} s_t^{(y)}.
\end{equation}

\section{Experimental Results}
\label{sec:experiments}

\subsection{Datasets}
To evaluate the proposed method, we used two publicly available online
handwriting datasets.
The first is Unipen~\cite{Guyon1994UNIPEN}, for which we used three subsets:
numerical digits (Unipen~1A), uppercase letters (Unipen~1B), and lowercase letters (Unipen~1C).
Each sample is represented as a sequence of 2D coordinates, and the sequence length was normalized to $T=50$ time steps.
Unipen~1A has $11{,}650$ training and $1{,}300$ test samples, while Unipen~1B and Unipen~1C have $11{,}063$ training and $1{,}235$ test samples.

The second is CASIA-OLHWDB 1.1~\cite{Liu2011CASIA}, an online handwritten Chinese character dataset with $3{,}755$ classes from $300$ writers.
For CASIA, we remove stroke-end and character-end sentinel coordinates, concatenate the strokes into a single 2D point sequence, and normalize the coordinates to $[-1,1]$ per sample and axis.
Unlike Unipen, CASIA samples are kept at their original variable lengths.
The test split contains $224{,}559$ samples.

\subsection{Architecture and Settings}
For the experiments, four classifiers are used.
\begin{itemize}
    \item \textbf{1D CNN-3}: A CNN with three convolutional blocks.
    Each block consists of a 1D convolution, BatchNorm~\cite{IoffeSzegedy2015BatchNorm}, a Rectified Linear Unit (ReLU), and max pooling.
    The first block uses 64 filters with kernel size 3 and stride 1, while the subsequent two blocks use 128 filters.
    The network is followed by two fully connected layers: one hidden layer with 512 units and dropout with probability 0.5, and one output layer.

    \item \textbf{1D CNN-4}: This network has the same architecture as 1D CNN-3, except that it contains four convolutional blocks instead of three.

    \item \textbf{BLSTM-2}: To test transferability to a different network structure, we use a Bidirectional LSTM (BLSTM)~\cite{schuster1997bidirectional}.
    Specifically, we use a two-layer BLSTM with 100 units per layer, following the recommendation of~\cite{reimers2017optimal}.

    \item \textbf{Transformer}: To test transferability to an attention-based architecture, we use an encoder-only Transformer~\cite{vaswani2017attention}.
    The model consists of a linear input projection, positional encoding, and six Transformer encoder blocks with four attention heads each.
    The encoder output is averaged over time and passed through a small MLP head to produce the class logits.
\end{itemize}

For CASIA-OLHWDB, we use analogous classifiers adapted to variable-length inputs and 3{,}755 classes: a 1D CNN with masked global average pooling in place of 1D CNN-3, together with BLSTM-2 and Transformer counterparts.
For CASIA-OLHWDB, we additionally train a second 1D CNN as a black-box target (channels 64→128→128→128, kernel size 3 throughout, 80 epochs, clean test accuracy 94.8\%), providing a within-CNN-family transfer target with a different filter configuration from the surrogate (64→128→256→256).

In both white-box and black-box settings, adversarial samples are generated using the gradients or salience of a 1D CNN surrogate model: 1D CNN-3 for Unipen and the analogous 1D CNN for CASIA-OLHWDB.
In the white-box setting, the generated adversarial samples are used to attack the same surrogate model.
In the black-box setting, they are transferred to different target models: 1D CNN-4, BLSTM-2, and Transformer for both datasets.
We only consider one-shot black-box attacks, where the attacker receives no feedback from the target classifiers.

For Unipen, all models are trained using the Adam optimizer~\cite{KingmaBa2015Adam} with an initial learning rate of $10^{-3}$ for 10,000 training iterations. For CASIA-OLHWDB, the CNNs and BLSTM are trained for 80 epochs and the Transformer for 100 epochs.

\subsection{Comparison Methods}

To evaluate the proposed method, we compare to seven well-established adversarial attacks: four classical attacks (FGSM, BIM, PGD, CW) and three transfer-oriented variants (MI-FGSM, NI-FGSM, 1D-TI-MI-FGSM).
The hyperparameters for each attack are as follows:
  \begin{itemize}
    \item \textbf{FGSM}~\cite{Goodfellow2015}: $\epsilon=0.1$ under the
    $L_\infty$ constraint.
    \item \textbf{BIM}~\cite{Kurakin2017AdversarialExamples}: $\epsilon=0.1$, $L_\infty$, step size $\epsilon_{\mathrm{iter}}=0.001$, and $I=600$ iterations.
    \item \textbf{PGD}~\cite{Madry2018}: same as BIM.
    \item \textbf{CW}~\cite{CarliniWagner2017}: $L_2$ constraint, $I=600$ iterations, learning rate $5\times10^{-3}$, 7 binary search steps, initial constant $5\times10^{-3}$, confidence $0$, and early stopping.
    \item \textbf{MI-FGSM}~\cite{Dong2018MIFGSM}: BIM with an accumulated momentum term over the input gradients to stabilize the update direction.
    Decay factor $\mu=1.0$; remaining settings as BIM.
    \item \textbf{NI-FGSM}~\cite{Lin2020NIFGSM}: MI-FGSM with the gradient evaluated at a Nesterov look-ahead point before each momentum update.
    Settings as MI-FGSM.
    \item \textbf{1D-TI-MI-FGSM}~\cite{Dong2019TIFGSM}: MI-FGSM in which the per-step gradient is convolved with a 1D Gaussian kernel
    (size $7$, $\sigma=3.0$) along the time axis. Remaining settings as MI-FGSM.
  \end{itemize}
  Since inputs are normalized to $[-1,1]$, adversarial examples are also clipped to $[-1,1]$.
  For CASIA-OLHWDB, $\epsilon=0.03$ and $I=400$ are used to account for the larger number of classes; the CW binary search steps are reduced to $3$.
  
\subsection{Results}
\begin{table}[t]
    \centering
    \scriptsize
    \setlength{\tabcolsep}{2.5pt}
    \renewcommand{\arraystretch}{0.78}
    \caption{Comparative evaluation of adversarial attacks and transferability.
    Accuracies (\%, $\downarrow$) on the surrogate (white-box) and on three one-shot black-box targets. Best per column, excluding ``No Attack'', is in bold; ``--'' indicates the target is not used.}
    \label{tab:results}
    \begin{tabular}{cl|c|ccc}
    \toprule
     & & White-Box & \multicolumn{3}{c}{One-Shot Black-Box} \\
    \cmidrule(lr){3-3} \cmidrule(lr){4-6}
    Dataset & Attack & 1D CNN-3 & 1D CNN-4 & BLSTM-2 & Transformer \\
    \midrule
    \multirow{9}{*}{Unipen 1A}
     & No Attack       & 98.9 & 98.8 & 98.8 & 98.2 \\
     & FGSM            & 73.8 & 82.8 & 95.6 & 92.9 \\
     & BIM             & 10.9 & 72.9 & 93.9 & 91.3 \\
     & PGD             & 12.4 & 74.2 & 95.1 & 92.0 \\
     & CW              & \textbf{0.69} & 93.9 & 98.2 & 97.1 \\
     & MI-FGSM         & 10.9 & 70.8 & 93.2 & 91.0 \\
     & NI-FGSM         & 9.1  & 70.5 & 93.7 & 91.0 \\
     & 1D-TI-MI-FGSM   & 67.5 & 71.2 & 80.6 & 83.3 \\
     & AITE (Ours)     & 25.5 & \textbf{58.2} & \textbf{70.2} & \textbf{68.8} \\
    \midrule
    \multirow{9}{*}{Unipen 1B}
     & No Attack       & 98.1 & 97.5 & 97.7 & 95.5 \\
     & FGSM            & 68.7 & 71.5 & 92.6 & 88.3 \\
     & BIM             & 11.2 & 75.4 & 91.1 & 85.8 \\
     & PGD             & 13.7 & 75.8 & 92.5 & 86.2 \\
     & CW              & \textbf{4.08} & 95.6 & 97.1 & 93.4 \\
     & MI-FGSM         & 11.2 & 71.3 & 91.1 & 84.5 \\
     & NI-FGSM         & 9.0  & 73.4 & 91.6 & 85.0 \\
     & 1D-TI-MI-FGSM   & 67.7 & 73.5 & 78.8 & 75.9 \\
     & AITE (Ours)     & 15.0 & \textbf{61.8} & \textbf{60.3} & \textbf{65.0} \\
    \midrule
    \multirow{9}{*}{Unipen 1C}
     & No Attack       & 97.1 & 97.7 & 96.5 & 94.8 \\
     & FGSM            & 64.5 & 79.4 & 90.8 & 85.2 \\
     & BIM             & 20.2 & 80.1 & 89.6 & 85.7 \\
     & PGD             & 30.4 & 85.8 & 93.6 & 89.1 \\
     & CW              & \textbf{5.96} & 96.9 & 96.1 & 93.8 \\
     & MI-FGSM         & 7.5  & 74.2 & 88.9 & 84.1 \\
     & NI-FGSM         & 6.2  & 75.6 & 88.7 & 85.3 \\
     & 1D-TI-MI-FGSM   & 55.6 & \textbf{68.9} & 76.7 & 73.4 \\
     & AITE (Ours)     & 18.4 & 70.5 & \textbf{59.7} & \textbf{62.0} \\
    \midrule
    \multirow{9}{*}{CASIA}
     & No Attack       & 95.0 & 94.8 & 96.5 & 96.3 \\
     & FGSM            & 65.2 & 78.5 & 90.1 & 90.7 \\
     & BIM             & 53.5 & 75.3 & 89.6 & 90.4 \\
     & PGD             & 53.8 & 75.2 & 89.7 & 90.4 \\
     & CW              & \textbf{0.71} & 74.4 & 90.5 & 93.0 \\
     & MI-FGSM         & 54.3 & 75.4 & 89.6 & 90.4 \\
     & NI-FGSM         & 53.6 & 75.2 & 89.6 & 90.4 \\
     & 1D-TI-MI-FGSM   & 85.3 & 89.9 & 94.2 & 93.5 \\
     & AITE (Ours)     & 29.2 & \textbf{52.5} & \textbf{53.6} & \textbf{47.9} \\
    \bottomrule
    \end{tabular}
    \renewcommand{\arraystretch}{1.0}
  \end{table}

Table~\ref{tab:results} reports recognition accuracy under adversarial attacks for both white-box and one-shot black-box settings across the three Unipen subsets and CASIA-OLHWDB.
In the white-box scenario, gradient-based attacks, specifically BIM, PGD, and CW, drastically reduce classification accuracy on the surrogate, with CW achieving the lowest accuracy on every dataset (0.69--5.96\% on Unipen and 0.71\% on CASIA).
MI-FGSM and NI-FGSM behave similarly to BIM in this setting, while 1D-TI-MI-FGSM is comparatively weaker on the surrogate because the smoothing of the input gradient sacrifices in-model optimality in exchange for transferability.

However, strong performance in the white-box setting does not necessarily carry over to the black-box scenario.
Although CW is highly effective when applied directly to the surrogate, it transfers poorly to unseen targets: its impact on the 1D CNN-4, BLSTM-2, and Transformer columns is consistently the smallest among all attacks (e.g., the recognition accuracy on the Transformer target stays above 93\%
on every dataset).
This is because gradient-based attacks rely on perturbations optimized with respect to the surrogate model's gradients, which may not align with the decision boundaries of an unseen target, and the emphasis of CW on minimizing the perturbation magnitude further limits its transferability.
MI-FGSM and NI-FGSM, which were designed for improved black-box transferability, do improve over plain BIM/PGD in some cases (e.g., Unipen 1A and 1B on 1D CNN-4), but they still leave the BLSTM-2 and Transformer targets largely intact, with accuracies above 84\% on every Unipen subset and CASIA target.
1D-TI-MI-FGSM is the strongest baseline overall and even slightly outperforms AITE on Unipen 1C against 1D CNN-4 (68.9 vs.\ 70.5).
However, it remains substantially weaker than AITE on the BLSTM-2 and Transformer targets across all datasets.

In contrast, the proposed AITE achieves the lowest recognition accuracy in almost every one-shot black-box column, including transfer to the attention-based Transformer target that was never used during attack
generation.
On CASIA-OLHWDB, AITE reduces the 1D CNN-4, BLSTM-2, and Transformer accuracies to 52.5\%, 53.6\%, and 47.9\%, respectively.
Notably, AITE transfers more effectively to the within-CNN-family target (CNN-4) than across architectures, consistent with the intuition that CNN-to-CNN perturbations share more decision-boundary structure.
The strongest additive baseline (1D-TI-MI-FGSM) only reduces the three black-box targets to 90.0\%, 94.2\%, and 93.5\%, respectively.
Because AITE selects insertion and deletion positions from salience rather than from raw input gradients, the resulting temporal edits target class-discriminative structure that tends to be shared across architectures, including CNN-to-BLSTM and CNN-to-Transformer transfers.

\subsection{Visual Similarity}

Attack success is usually quantified by the change in the model's prediction. However, for handwriting recognition, misclassification alone does not characterize the practical relevance of an attack: handwritten inputs are interpreted by humans, and visibly distorted adversarial samples are unlikely to represent plausible handwriting.
We therefore also assess the visual similarity between original and adversarial samples.

\begin{table}[t]
  \centering
  \scriptsize
  \setlength{\tabcolsep}{3pt}
  \renewcommand{\arraystretch}{0.85}
  \caption{Visual similarity between adversarial and original samples.
  The best value for each metric and dataset is in bold.}
  \label{tab:similarity}
  \begin{tabular}{clcccccccc}
    \toprule
    Dataset & Metric
    & FGSM & BIM & PGD & CW & MI-FGSM & NI-FGSM & 1D-TI-MI-FGSM & AITE \\
    \midrule
    \multirow{2}{*}{Unipen 1A}
     & $L_2$ $\downarrow$
     & 22.0 & 20.9 & 21.0 & \textbf{13.8} & 21.8 & 21.4 & 22.8 & 14.3 \\
     & SSIM $\uparrow$
     & 0.550 & 0.577 & 0.574 & \textbf{0.760} & 0.554 & 0.563 & 0.518 & 0.746 \\
    \midrule
    \multirow{2}{*}{Unipen 1B}
     & $L_2$ $\downarrow$
     & 23.5 & 22.4 & 22.5 & 15.0 & 23.2 & 22.9 & 24.3 & \textbf{11.7} \\
     & SSIM $\uparrow$
     & 0.497 & 0.525 & 0.524 & 0.725 & 0.502 & 0.512 & 0.466 & \textbf{0.808} \\
    \midrule
    \multirow{2}{*}{Unipen 1C}
     & $L_2$ $\downarrow$
     & 22.1 & 19.2 & 19.4 & 13.1 & 21.7 & 21.4 & 22.7 & \textbf{11.7} \\
     & SSIM $\uparrow$
     & 0.551 & 0.627 & 0.624 & 0.775 & 0.558 & 0.567 & 0.527 & \textbf{0.809} \\
    \midrule
    \multirow{2}{*}{CASIA}
     & $L_2$ $\downarrow$
     & 17.8 & 17.6 & 17.6 & \textbf{11.1} & 17.6 & 17.6 & 18.3 & 15.9 \\
     & SSIM $\uparrow$
     & 0.585 & 0.593 & 0.594 & \textbf{0.812} & 0.592 & 0.593 & 0.564 & 0.618 \\
    \bottomrule
  \end{tabular}
  \renewcommand{\arraystretch}{1.0}
\end{table}

To this end, we compare the visual similarity of the rendered characters. Table~\ref{tab:similarity} reports quantitative results across all attacks. In this table, the online handwritten digits and characters are rendered into $64\times64$ grayscale images with black strokes and no points.
For the visual similarity measures, we use the pixel-space $L_2$ distance and the Structural Similarity Index Measure~(SSIM) between the original samples and the adversarial samples.
$L_2$ measures the strict pixel-wise differences between the images, whereas SSIM emphasizes structural consistency, which better reflects human visual perception.

\begin{figure}[!t]
\centering
\includegraphics[width=\linewidth]{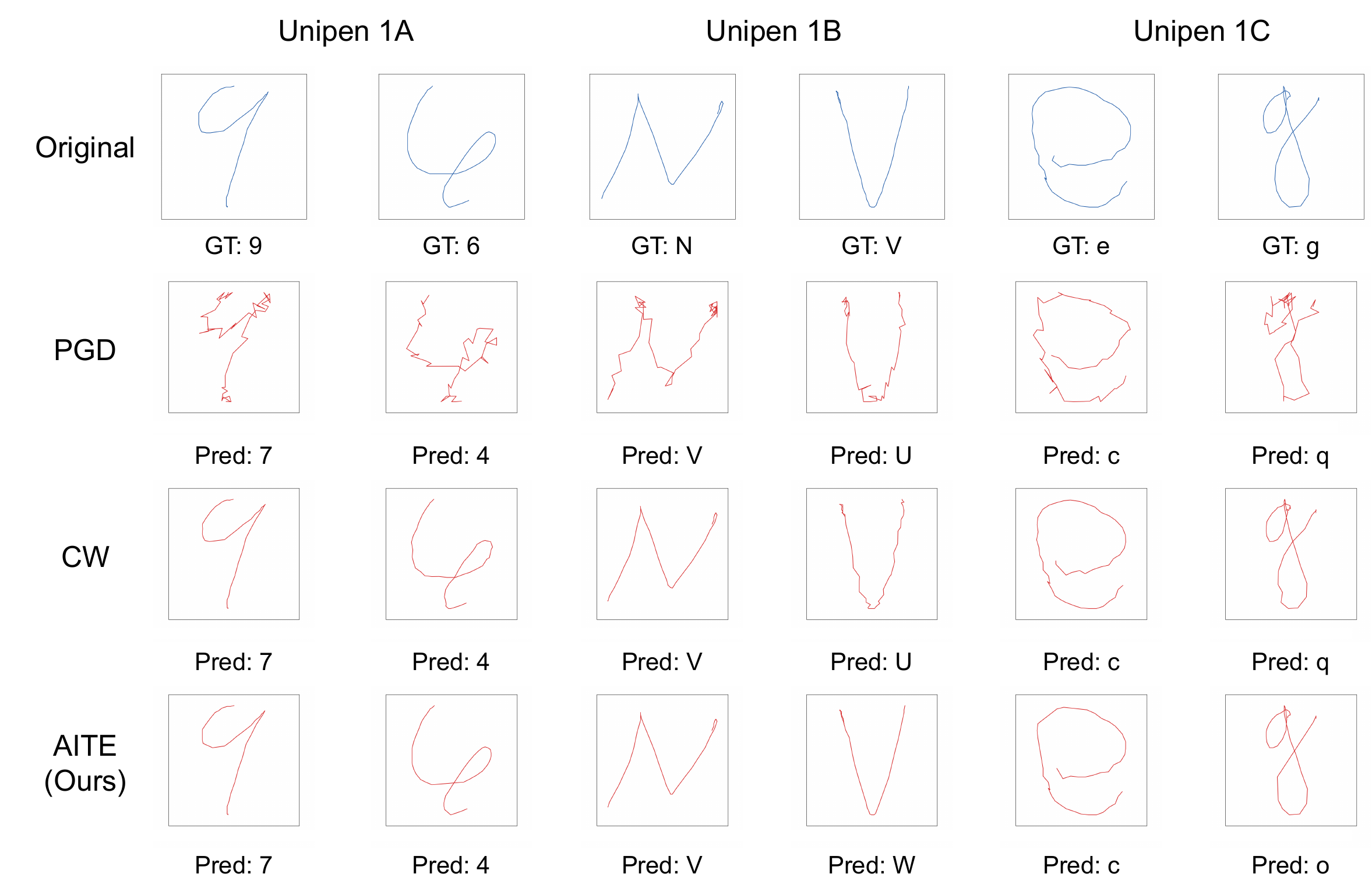}
\caption{Qualitative comparison of adversarial handwriting trajectories on Unipen (1A/1B/1C) for representative attacks. Rows show the original sample and adversarial samples generated by PGD, CW, and the proposed AITE. Ground-truth labels (GT) and model predictions (Pred) are shown below each example.}
\label{fig:adversarial_unipen_samples}
\end{figure}

FGSM, BIM, PGD, MI-FGSM, NI-FGSM, and 1D-TI-MI-FGSM consistently produce large $L_2$ distortions and low SSIM scores across all datasets.
As shown in Fig.~\ref{fig:adversarial_unipen_samples}, this reflects the large amount of jittering in the generated strokes of the characters.
In the rendered examples, these attacks introduce visible jitter and local stroke irregularities.
The transfer-oriented variants (MI-FGSM, NI-FGSM, and 1D-TI-MI-FGSM) do not qualitatively change this behavior: they share the additive $L_\infty$ formulation of BIM/PGD, so the rendered trajectories exhibit similar jittering and remain easily detectable.
In contrast, the perturbations in CW are minimal, but as discussed above, this adversely affects black-box transferability because the minimal perturbations only exploit surrogate-specific decision boundaries.

In contrast, AITE preserves the overall structure of the handwriting while still being effective under black-box transfer.
Although online handwriting is represented as a time series, temporal-domain edits such as point insertion or deletion alter the temporal structure of the sequence while leaving the rendered spatial trace largely unchanged, and are typically less perceptible than the spatial jitter introduced by additive perturbations.

\subsection{Gradient Versus Salience for Temporal Edit Selection}

As a baseline ablation, we replace the Grad-CAM salience with gradient-based importance: $s^{grad}_t = \lVert\nabla_{\mathbf{x}_t}\mathcal{L}(f_\theta(\mathbf{X}), y)\rVert_2$, where larger values indicate higher sensitivity of the loss to time step $t$.

\begin{table}[t]
\centering
\setlength{\tabcolsep}{4pt}
\caption{Comparative evaluation of AITE selection strategies and transferability. For the white-box experiment, the adversarial samples are generated and tested on 1D CNN-3. For the black-box experiments, the adversarial examples are generated from 1D CNN-3 and tested on 1D CNN-4 and BLSTM-2 models. The best of each dataset is in bold.}
\label{tab:gradient}
\begin{tabular}{cl|cc|c|cc}
\toprule
 & & \multicolumn{2}{c|}{Visual} & White-Box & \multicolumn{2}{c}{One-Shot Black-Box} \\
 & & \multicolumn{2}{c|}{Similarity} & 1D CNN-3 & 1D CNN-4 & BLSTM-2 \\
Dataset & Selection & $L_2$ $\downarrow$ & SSIM $\uparrow$ & Acc (\%) $\downarrow$ & Acc (\%) $\downarrow$  & Acc (\%) $\downarrow$ \\
\midrule
\multirow{2}{*}{Unipen 1A} 
& Gradient & \textbf{9.92} & \textbf{0.849} & 64.6 & 76.8 & 88.8 \\
& Salience & 14.3 & 0.746 & \textbf{25.5} & \textbf{58.2} & \textbf{70.2} \\

\midrule
\multirow{2}{*}{Unipen 1B}
& Gradient & \textbf{9.81} & \textbf{0.846} & 50.1 & 64.8 & 72.8 \\
& Salience & 11.7 & 0.808 & \textbf{15.0} & \textbf{61.8} & \textbf{60.3} \\
\midrule
\multirow{2}{*}{Unipen 1C}
& Gradient & \textbf{9.46} & \textbf{0.855} & 53.9 & 82.1 & 81.6 \\
& Salience & 11.7 & 0.809 & \textbf{18.4} & \textbf{70.5} & \textbf{59.7} \\
\bottomrule
\end{tabular}
\end{table}

A comparison of the results is shown in Table~\ref{tab:gradient}.
In the results, the gradient-based selection creates more similar adversarial samples but has a smaller effect on the accuracy.
The reason for this is because the edited time steps tend to be adjacent to each other in salient regions. 
Conversely, time steps selected by gradient magnitudes tend to be distributed more equally over the character, allowing for less disruptive edits.
However, both the white-box and black-box attacks with gradient-based selection are not as effective as salience-based selection.
This suggests salience is more suitable than gradients for temporal editing.

\section{Examination of the Temporal Edits}

\subsection{Single-Iteration Adversarial Samples}

\begin{figure}[t]
  \centering
  \includegraphics[width=0.95\linewidth]{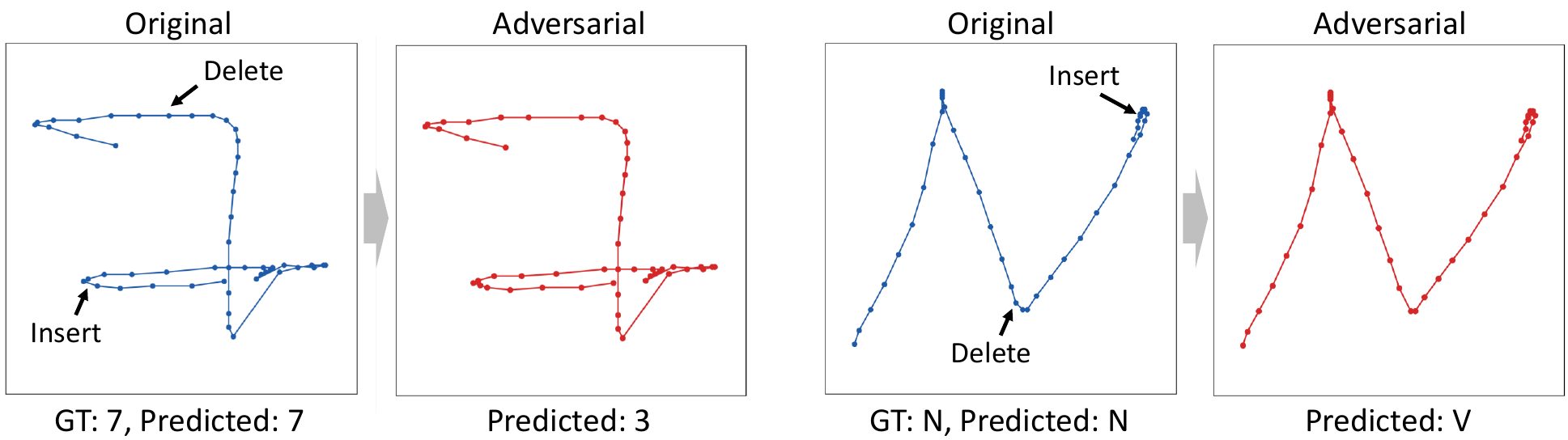}
  \caption{Example of adversarial samples that only required a single iteration of AITE to change the class. The temporal insertion and deletion are indicated with arrows.}
  \label{fig:one-step}
\end{figure}

An interesting observation from our experiments is that some handwriting samples can be misclassified after only a single iteration of temporal editing. 
In other words, the class of a character is able to be changed with the removal of a single time step and the addition of a single time step.
As demonstrated in Fig.~\ref{fig:one-step}, these cases are nearly identical to the original samples.
These cases are of interest because they show samples whose correct classification relies on a single time step.

\subsection{Class Sensitivity to Temporal Edits}

\begin{figure}[t]
  \centering
  \begin{subfigure}[b]{0.32\textwidth}
        \centering
        \includegraphics[width=\textwidth,clip,trim=0 0 10cm 8.5cm]{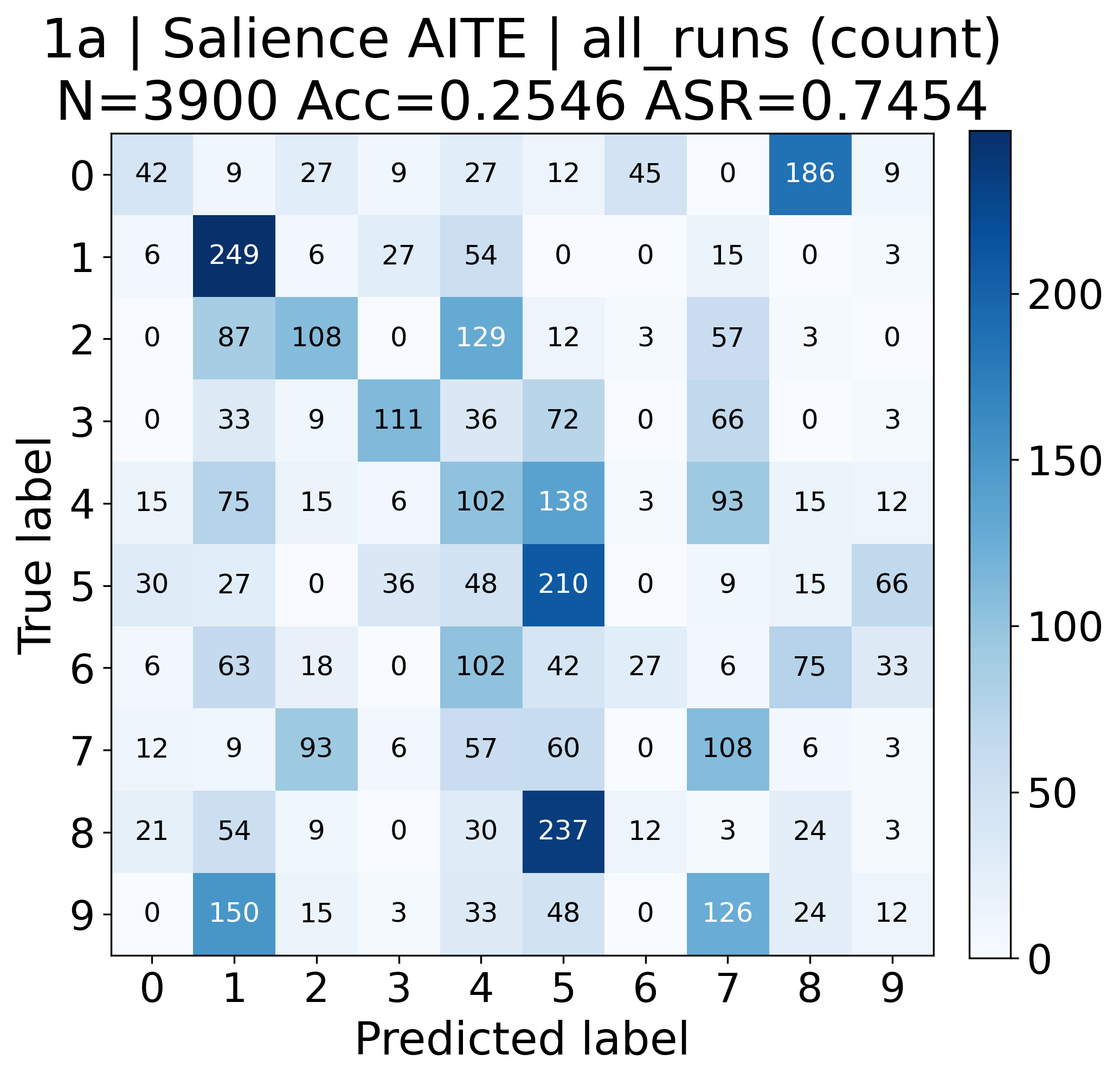}
        \caption{AITE (Ours)}
        \label{fig:sub1}
    \end{subfigure}
    \hfill
    \begin{subfigure}[b]{0.32\textwidth}
        \centering
        \includegraphics[width=\textwidth,clip,trim=0 0 10cm 8.5cm]{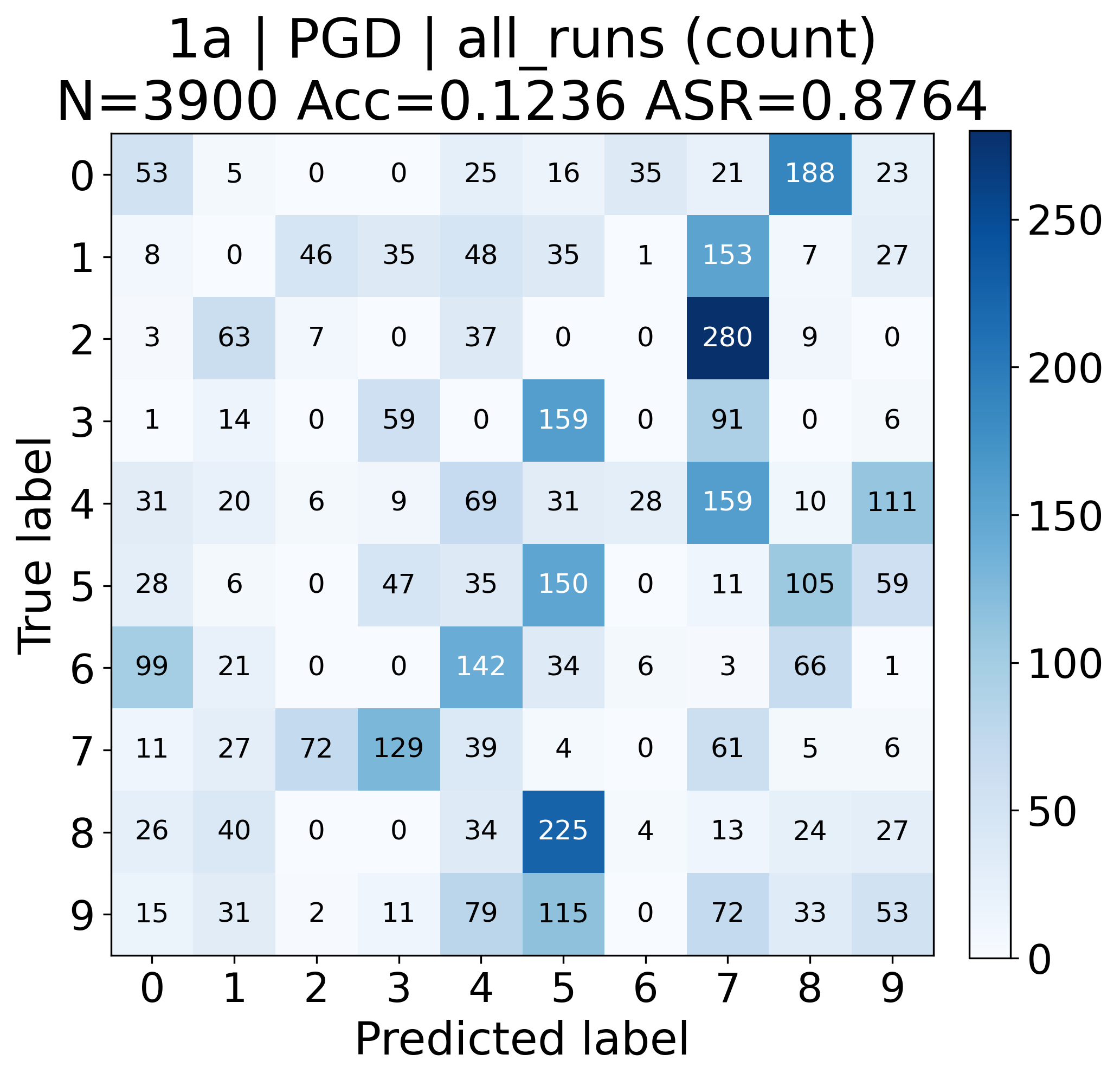}
        \caption{PGD}
        \label{fig:sub2}
    \end{subfigure}
    \hfill
    \begin{subfigure}[b]{0.32\textwidth}
        \centering
        \includegraphics[width=\textwidth,clip,trim=0 0 10cm 8.5cm]{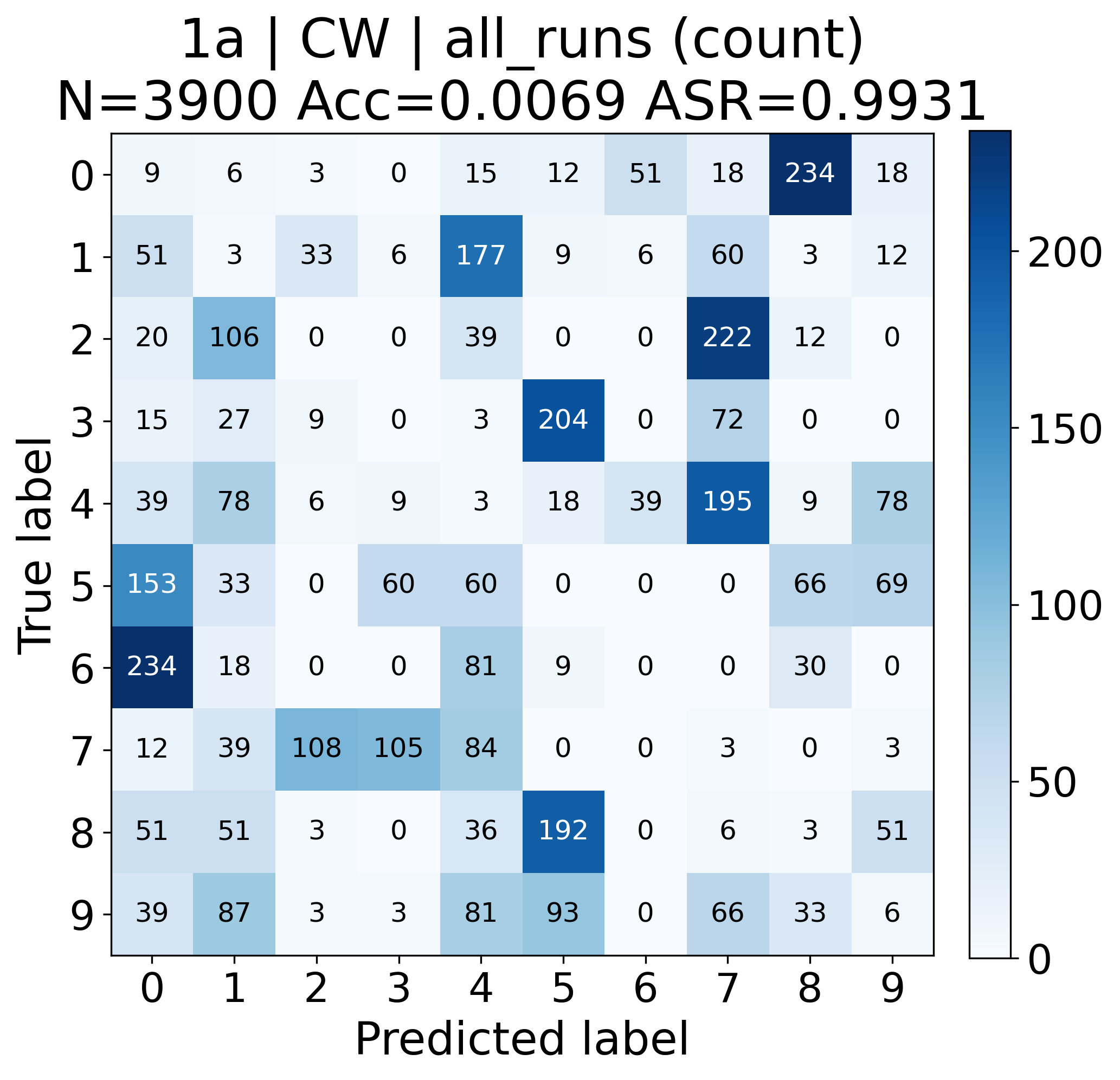}
        \caption{CW}
        \label{fig:sub3}
    \end{subfigure}
  \caption{Comparison of the confusion matrices on Unipen 1A under different attacks. Darker cells indicate larger counts.}
  \label{fig:confusion}
\end{figure}

Fig.~\ref{fig:confusion} compares the confusion matrices on Unipen 1A under AITE, PGD, and CW.
PGD and CW, being gradient-based, produce similar confusion patterns, whereas AITE differs substantially: for instance, ``2''$\to$``4'' under AITE versus ``2''$\to$``7'' under the gradient methods, suggesting that
``2''s are temporally closer to ``4''s but spatially closer to ``7''s.
``9''s also tend to transition to ``1''s or ``7''s under AITE, while a few classes behave similarly across methods (e.g., ``0''$\to$``8'' and ``8''$\to$``5'').

The matrices also reveal class-wise susceptibility: ``1'' and ``5'' are the hardest digits to attack with AITE because their single-stroke shapes leave little room for class change without spatial distortion, while the more complex ``6'', ``8'', and ``9'' are the easiest. This is consistent with the stronger attack effectiveness on the more complex Unipen 1B and
1C subsets in Table~\ref{tab:results}.

\section{Conclusion}
This paper examined adversarial attacks on online handwriting recognition from a temporal perspective.
By formulating adversarial example generation as a sequence editing problem, the proposed approach modifies handwriting trajectories through point insertion and deletion guided by temporal salience, rather than through additive spatial perturbations. 
Experimental evaluations under both white-box and one-shot black-box settings demonstrated that temporal editing produces adversarial examples that preserve the structure and smoothness of natural handwritten characters while remaining effective across different model architectures.
Future work will extend this preliminary study to word- and sentence-level online handwriting recognition. Human evaluation should also be conducted to assess the perceived naturalness of adversarial handwriting, and defenses against temporal editing attacks should be investigated.
\subsubsection*{Acknowledgements}
This research was partially funded by JSPS Grant Number 25K22845 and JST CRONOS-JPMJCS24K4.

\let\oldthebibliography\thebibliography
\let\endoldthebibliography\endthebibliography
\renewenvironment{thebibliography}[1]{%
  \begin{oldthebibliography}{#1}%
    \setlength{\itemsep}{0pt}%
    \setlength{\parskip}{0pt}%
}{\end{oldthebibliography}}
\enlargethispage{10\baselineskip}
\bibliographystyle{splncs04}
\bibliography{mybibliography}

@inproceedings{Liu2011CASIA,
    author    = {Liu, Cheng-Lin and Yin, Fei and Wang, Da-Han and Wang, Qiu-Feng},
    title     = {{CASIA} Online and Offline {C}hinese Handwriting Databases},
    booktitle = {Int. Conf. Doc. Anal. Recognit. (ICDAR)},
    year      = {2011}
}

@article{Graves2009TPAMIHandwriting,
  author  = {Graves, Alex and Liwicki, Marcus and Fern{\'a}ndez, Santiago and Bertolami, Roman and Bunke, Horst and Schmidhuber, J{\"u}rgen},
  title   = {A Novel Connectionist System for Unconstrained Handwriting Recognition},
  journal = {IEEE Trans. Pattern Anal. Mach. Intell. (TPAMI)},
  volume  = {31},
  number  = {5},
  pages   = {855--868},
  year    = {2009}
}

@inproceedings{Guyon1994UNIPEN,
  author    = {Guyon, Isabelle M. and Schomaker, Lambert and Plamondon, R{\'e}jean and Liberman, Mark and Janet, Stan},
  title     = {UNIPEN project of on-line data exchange and recognizer benchmarks},
  booktitle = {Int. Conf. Pattern Recognit. (ICPR)},
  year      = {1994},
  volume    = {2},
  pages     = {29--33},
}

@inproceedings{Goodfellow2015,
  title     = {Explaining and Harnessing Adversarial Examples},
  author    = {Goodfellow, Ian J. and Shlens, Jonathon and Szegedy, Christian},
  booktitle = {Int. Conf. Learn. Represent. (ICLR)},
  year      = {2015}
}

@inproceedings{Madry2018,
  title     = {Towards Deep Learning Models Resistant to Adversarial Attacks},
  author    = {Madry, Aleksander and Makelov, Aleksandar and Schmidt, Ludwig and Tsipras, Dimitris and Vladu, Adrian},
  booktitle = {Int. Conf. Learn. Represent. (ICLR)},
  year      = {2018}
}

@inproceedings{Kurakin2017AdversarialExamples,
  author    = {Kurakin, Alexey and Goodfellow, Ian and Bengio, Samy},
  title     = {Adversarial Examples in the Physical World},
  booktitle = {Workshop Int. Conf. Learn. Represent. (ICLR)},
  year      = {2017}
}

@inproceedings{CarliniWagner2017,
  title     = {Towards Evaluating the Robustness of Neural Networks},
  author    = {Carlini, Nicholas and Wagner, David},
  booktitle = {SP},
  pages     = {39--57},
  year      = {2017},
  publisher = {IEEE}
}

@inproceedings{Carlini2016HiddenCommands,
  author    = {Carlini, Nicholas and Mishra, Pratyush and Vaidya, Tavish and Zhang, Yuankai and Sherr, Micah and Shields, Clay and Wagner, David and Zhou, Wenchao},
  title     = {Hidden Voice Commands},
  booktitle = {USENIX},
  year      = {2016},
  publisher = {USENIX Association},
}

@inproceedings{Fawaz2019AdversarialTSC,
  title={Adversarial attacks on deep neural networks for time series classification},
  author={Fawaz, Hassan Ismail and Forestier, Germain and Weber, Jonathan and Idoumghar, Lhassane and Muller, Pierre-Alain},
  booktitle={Int. Joint Conf. Neural Netw. (IJCNN)},
  year={2019},
  organization={IEEE}
}

@article{reimers2017optimal,
  title={Optimal hyperparameters for deep lstm-networks for sequence labeling tasks},
  author={Reimers, Nils and Gurevych, Iryna},
  journal={arXiv preprint arXiv:1707.06799},
  year={2017}
}

@article{schuster1997bidirectional,
  title={Bidirectional recurrent neural networks},
  author={Schuster, Mike and Paliwal, Kuldip K},
  journal={IEEE Trans. Signal Process.},
  volume={45},
  number={11},
  pages={2673--2681},
  year={1997},
  publisher={IEEE}
}

@inproceedings{Dong2018MIFGSM,
    title     = {Boosting Adversarial Attacks with Momentum},
    author    = {Dong, Yinpeng and Liao, Fangzhou and Pang, Tianyu and Su, Hang and Zhu, Jun and Hu, Xiaolin and Li, Jianguo},
    booktitle = {IEEE/CVF Conf. Comput. Vis. Pattern Recognit. (CVPR)},
    year      = {2018}
  }

@inproceedings{Lin2020NIFGSM,
    title     = {Nesterov Accelerated Gradient and Scale Invariance for Adversarial Attacks},
    author    = {Lin, Jiadong and Song, Chuanbiao and He, Kun and Wang, Liwei and Hopcroft, John E.},
    booktitle = {Int. Conf. Learn. Represent. (ICLR)},
    year      = {2020}
  }

@inproceedings{Dong2019TIFGSM,
    title     = {Evading Defenses to Transferable Adversarial Examples by Translation-Invariant Attacks},
    author    = {Dong, Yinpeng and Pang, Tianyu and Su, Hang and Zhu, Jun},
    booktitle = {IEEE/CVF Conf. Comput. Vis. Pattern Recognit. (CVPR)},
    year      = {2019}
  }

@inproceedings{Selvaraju2017GradCAM,
  author    = {Selvaraju, Ramprasaath R. and Cogswell, Michael and Das, Abhishek and Vedantam, Ramakrishna and Parikh, Devi and Batra, Dhruv},
  title     = {Grad-CAM: Visual Explanations from Deep Networks via Gradient-Based Localization},
  booktitle = {Int. Conf. Comput. Vis. (ICCV)},
  pages     = {618--626},
  year      = {2017},
  publisher = {IEEE}
}

@inproceedings{KingmaBa2015Adam,
  title={Adam: A method for stochastic optimization},
  author={Kingma, Diederik P and Ba, Jimmy},
  booktitle = {Int. Conf. Learn. Represent. (ICLR)},
  year={2015}
}

@inproceedings{IoffeSzegedy2015BatchNorm,
  author    = {Ioffe, Sergey and Szegedy, Christian},
  title     = {Batch Normalization: Accelerating Deep Network Training by Reducing Internal Covariate Shift},
  booktitle = {Int. Conf. Mach. Learn.},
  pages     = {448--456},
  year      = {2015},
}

@article{Papernot2016Transferability,
  title={Transferability in machine learning: from phenomena to black-box attacks using adversarial samples},
  author={Papernot, Nicolas and McDaniel, Patrick and Goodfellow, Ian},
  journal={arXiv preprint arXiv:1605.07277},
  year={2016}
}

@article{qu2025end,
  title={End-to-end multi-scale attention convolutional recurrent network for online handwritten Chinese text recognition},
  author={Qu, Xiwen and Wu, Zhihong},
  journal={Expert Syst. Appl.},
  pages={127626},
  year={2025},
  publisher={Elsevier}
}

@article{karim2020adversarial,
  title={Adversarial attacks on time series},
  author={Karim, Fazle and Majumdar, Somshubra and Darabi, Houshang},
  journal={IEEE Trans. Pattern Anal. Mach. Intell. (TPAMI)},
  volume={43},
  number={10},
  pages={3309--3320},
  year={2021},
  publisher={IEEE}
}

@article{xu2020adversarial,
  title={Adversarial attacks and defenses in images, graphs and text: A review},
  author={Xu, Han and Ma, Yao and Liu, Hao-Chen and Deb, Debayan and Liu, Hui and Tang, Ji-Liang and Jain, Anil K},
  journal={Int. J. Autom. Comput.},
  volume={17},
  number={2},
  pages={151--178},
  year={2020},
  publisher={Springer}
}

@article{costa2024deep,
  title={How deep learning sees the world: A survey on adversarial attacks \& defenses},
  author={Costa, Joana C and Roxo, Tiago and Proen{\c{c}}a, Hugo and Inacio, Pedro Ricardo Morais},
  journal={IEEE Access},
  volume={12},
  pages={61113--61136},
  year={2024},
  publisher={IEEE}
}

@inproceedings{vaswani2017attention,
  title={Attention is all you need},
  author={Vaswani, Ashish and Shazeer, Noam and Parmar, Niki and Uszkoreit, Jakob and Jones, Llion and Gomez, Aidan N and Kaiser, {\L}ukasz and Polosukhin, Illia},
  booktitle={Adv. Neural Inf. Process. Syst. (NeurIPS)},
  volume={30},
  year={2017}
}

@incollection{rumelhart1985learning,
  title={{Learning internal representations by error propagation}},
  author={Rumelhart, David E. and Hinton, Geoffrey E. and Williams, Ronald J.},
  booktitle={{Neurocomput.: Found. Res.}},
  editor={Anderson, James A. and Rosenfeld, Edward},
  publisher={{MIT Press}},
  pages={673--695},
  year={1988}
}

@article{lang1990time,
  title={A time-delay neural network architecture for isolated word recognition},
  author={Lang, Kevin J and Waibel, Alex H and Hinton, Geoffrey E},
  journal={Neural Netw.},
  volume={3},
  number={1},
  pages={23--43},
  year={1990},
  publisher={Elsevier}
}

@inproceedings{liu2025col,
  title={Col-OLHTR: A Novel Framework for Multimodal Online Handwritten Text Recognition},
  author={Liu, Chenyu and Hu, Jinshui and Yin, Baocai and Pan, Jia and Yin, Bing and Du, Jun and Liu, Qingfeng},
  booktitle={IEEE Int. Conf. Acoust. Speech Signal Process.},
  pages={1--5},
  year={2025},
}

@article{ghosh2022advances,
  title={Advances in online handwritten recognition in the last decades},
  author={Ghosh, Trishita and Sen, Shibaprasad and Obaidullah, Sk Md and Santosh, KC and Roy, Kaushik and Pal, Umapada},
  journal={Comput. Sci. Rev.},
  volume={46},
  year={2022},
  publisher={Elsevier}
}

@article{iwana2020time,
  title={Time series classification using local distance-based features in multi-modal fusion networks},
  author={Iwana, Brian Kenji and Uchida, Seiichi},
  journal={Pattern Recognit.},
  volume={97},
  pages={107024},
  year={2020}
}

@inproceedings{guo2024white,
  title={A white-box false positive adversarial attack method on contrastive loss based offline handwritten signature verification models},
  author={Guo, Zhongliang and Li, Weiye and Qian, Yifei and Arandjelovic, Ognjen and Fang, Lei},
  booktitle={Int. Conf. Artif. Intell. Stat. (AISTATS)},
  pages={901--909},
  year={2024},
}

@inproceedings{lopresti2005effectiveness,
  title={The effectiveness of generative attacks on an online handwriting biometric},
  author={Lopresti, Daniel P and Raim, Jarret D},
  booktitle={Int. Conf. Audio Video-Based Biometric Person Authent. (AVBPA)},
  pages={1090--1099},
  year={2005},
}

@inproceedings{yamashita2024test,
  title={Test Time Augmentation as a Defense Against Adversarial Attacks on Online Handwriting},
  author={Yamashita, Yoh and Iwana, Brian Kenji},
  booktitle={Int. Conf. Doc. Anal. Recognit. (ICDAR)},
  pages={156--172},
  year={2024},
}

@inproceedings{huynh2025adversarial,
  title={Adversarial Robustness Evaluation of a Vietnamese Handwriting OCR System},
  author={Huynh, Thai Bao and Phan, Quan Minh and Duong, Viet-Hang and Luong, Ngoc Hoang},
  booktitle={Int. Conf. Multimedia Anal. Pattern Recognit. (MAPR)},
  pages={1--6},
  year={2025},
}

@article{li2021black,
  title={Black-box attack against handwritten signature verification with region-restricted adversarial perturbations},
  author={Li, Haoyang and Li, Heng and Zhang, Hansong and Yuan, Wei},
  journal={Pattern Recognit.},
  volume={111},
  pages={107689},
  year={2021},
  publisher={Elsevier}
}

@article{hafemann2019characterizing,
  title={Characterizing and evaluating adversarial examples for offline handwritten signature verification},
  author={Hafemann, Luiz G and Sabourin, Robert and Oliveira, Luiz S},
  journal={IEEE Trans. Inf. Forensics Secur.},
  volume={14},
  number={8},
  pages={2153--2166},
  year={2019},
  publisher={IEEE}
}

@article{shi2025generative,
  title={A generative adversarial network-based client-level handwriting forgery attack in federated learning scenario},
  author={Shi, Lei and Wu, Han and Ding, Xu and Xu, Hao and Pan, Sinan},
  journal={Expert Sys.},
  volume={42},
  number={2},
  pages={e13676},
  year={2025},
  publisher={Wiley Online Library}
}

@article{zheng2025blackboxbench,
  title={Blackboxbench: A comprehensive benchmark of black-box adversarial attacks},
  author={Zheng, Meixi and Yan, Xuanchen and Zhu, Zihao and Chen, Hongrui and Wu, Baoyuan},
  journal={IEEE Trans. Pattern Anal. Mach. Intell. (TPAMI)},
  year={2025},
  publisher={IEEE},
  pages={7867-–7885},
  volume={47},
  number={9}
}

@inproceedings{jiang2022adversarial,
  title={Adversarial attack and defence on handwritten Chinese character recognition},
  author={Jiang, Guoteng and Qian, Zhuang and Wang, Qiu-Feng and Wei, Yan and Huang, Kaizhu},
  booktitle={J. Phys. Conf. Ser.},
  volume={2278},
  number={1},
  pages={012023},
  year={2022},
}

@inproceedings{pialla2022smooth,
  title={Smooth perturbations for time series adversarial attacks},
  author={Pialla, Gautier and Fawaz, Hassan Ismail and Devanne, Maxime and Weber, Jonathan and Idoumghar, Lhassane and Muller, Pierre-Alain and Bergmeir, Christoph and Schmidt, Daniel and Webb, Geoffrey and Forestier, Germain},
  booktitle={Pac.-Asia Conf. Knowl. Discovery Data Min. (PAKDD)},
  pages={485--496},
  year={2022},
  organization={Springer}
}

@inproceedings{rathore2020untargeted,
  title={Untargeted, targeted and universal adversarial attacks and defenses on time series},
  author={Rathore, Pradeep and Basak, Arghya and Nistala, Sri Harsha and Runkana, Venkataramana},
  booktitle={Int. Joint Conf. Neural Netw. (IJCNN)},
  pages={1--8},
  year={2020},
}

@inproceedings{ding2023black,
  title={Black-box adversarial attack on time series classification},
  author={Ding, Daizong and Zhang, Mi and Feng, Fuli and Huang, Yuanmin and Jiang, Erling and Yang, Min},
  booktitle={AAAI Conf. Artif. Intell. (AAAI)},
  volume={37},
  number={6},
  pages={7358--7368},
  year={2023}
}

@article{yang2022tsadv,
  title={TSadv: Black-box adversarial attack on time series with local perturbations},
  author={Yang, Wenbo and Yuan, Jidong and Wang, Xiaokang and Zhao, Peixiang},
  journal={Eng. Appl. Artif. Intell.},
  volume={114},
  pages={105218},
  year={2022},
  publisher={Elsevier}
}

@article{szegedy2013intriguing,
  title={Intriguing properties of neural networks},
  author={Szegedy, Christian and Zaremba, Wojciech and Sutskever, Ilya and Bruna, Joan and Erhan, Dumitru and Goodfellow, Ian and Fergus, Rob},
  journal={arXiv preprint arXiv:1312.6199},
  year={2013}
}

\end{document}